\documentclass[journal]{IEEEtran}
\usepackage{amsmath,amsfonts}
\usepackage{algorithmic}
\usepackage{algorithm}
\usepackage{array}
\usepackage{textcomp}
\usepackage{stfloats}
\usepackage{url}
\usepackage{verbatim}
\usepackage{graphicx}
\usepackage{cite}
\usepackage{graphics, epsfig}      
\usepackage{graphicx}              
\usepackage{verbatim}              
\usepackage{textcomp}              
\usepackage{url}                   
\usepackage{booktabs}              
\usepackage{array}                 
\usepackage{multirow, makecell}    
\usepackage{colortbl, hhline, diagbox} 
\usepackage{adjustbox}             

\usepackage{amsmath, amssymb}      
\usepackage{gensymb}               
\usepackage{siunitx}               
\usepackage{chemformula}           

\usepackage{mathptmx}              
\usepackage{times}                 
\usepackage{ulem}                  
\usepackage{soul}                  
\usepackage{xcolor}                
\usepackage{wasysym, pifont}       
\usepackage{makecell}

\usepackage{tikz}                  
\usepackage{svg}                   
\usepackage{subcaption}            
\usepackage[font=small]{caption}   
\usepackage[outline]{contour}      
\usepackage{cuted}                
\usepackage{tabularx}

\usepackage{hyperref}              
\usepackage{cite}                  
\usepackage{multibib}              
\newcites{R}{References for Response}

\usepackage{lipsum}                
\usepackage{comment}               

\usepackage{rotating}

\newcommand{\teaserfig}{%
  \vspace{2em} 
  \centering
    \includegraphics[width=.95\textwidth]{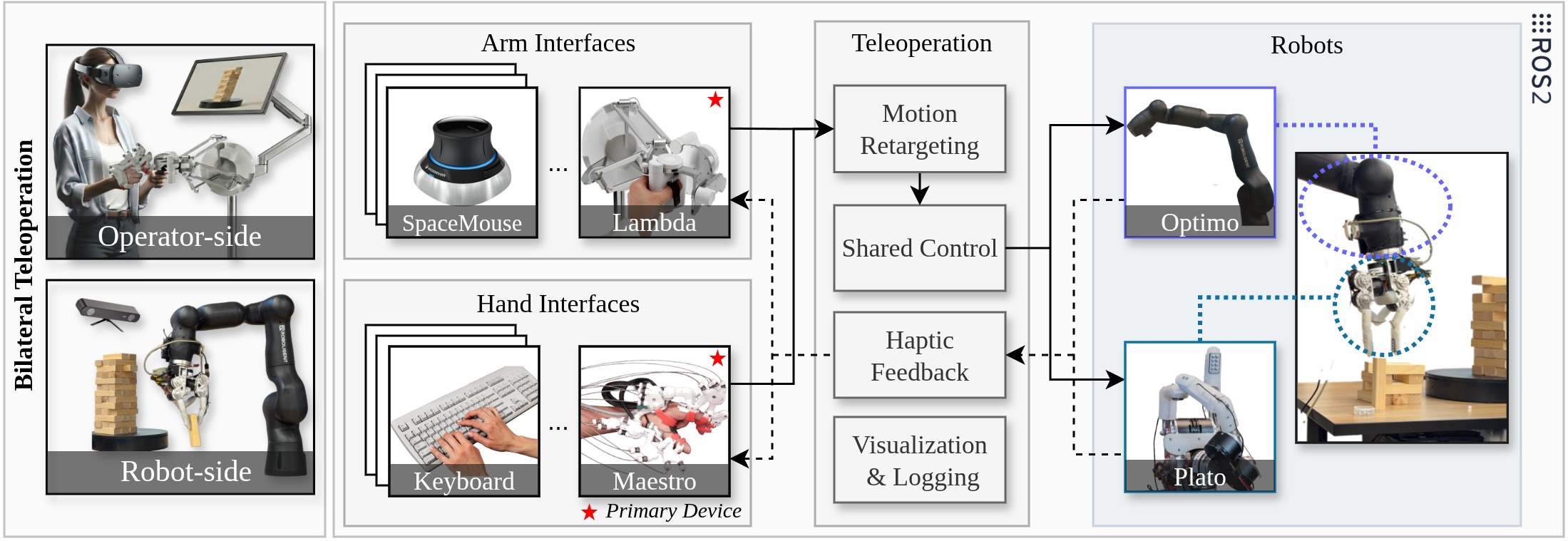}
    \captionsetup{type=figure}
    \setcounter{figure}{0} 
    \caption{(Left) Bilateral teleoperation framework with operator-side and robot-side stations. (Right) System architecture with modular arm--hand input interfaces, teleoperation control, and robot execution.}
    \label{fig:system_overview}
    \vspace{-2em}
}

\makeatletter
\apptocmd{\@maketitle}{\teaserfig}{}{}
\makeatother

\begin{document}

\title{A Bilateral Teleoperation Framework for Dexterous Manipulation}

\author{Stefano Dalla Gasperina$^{1,*,\dagger}$, 
Dong Ho Kang$^{1,\dagger}$,
Haiyun Zhang$^{1,\dagger}$,
Aldo Galvan$^{1,\dagger}$, 
Job D. Ramirez$^{1}$, \\
Aaron Kim$^{1}$,
Mark Helwig$^{1}$,
Kazuto Yokoyama$^{2}$, 
Takahisa Ueno$^{2}$,
Tetsuya Narita$^{2}$, \\
Ann Majewicz-Fey$^{1}$,  
Ashish D. Deshpande$^{1,3}$,
and Luis Sentis$^{1}$%

\thanks{This work is funded and supported by Sony Corporation Inc., Tokyo, Japan.}%
\thanks{$^1$The University of Texas at Austin, Austin, TX, USA}
\thanks{$^2$Sony Group Corporation, Tokyo, Japan}
\thanks{$^3$Meta Reality Labs Research, Redmond, WA, USA}%
\thanks{$^*$ Corresponding author}
\thanks{$^\dagger$ These authors contributed equally to this work.}
}

\markboth{IEEE Robotics and Automation Practice (RA-P) (SUBMITTED)}%
{Shell \MakeLowercase{\textit{et al.}}: A Sample Article Using IEEEtran.cls for IEEE Journals}



\maketitle
\setcounter{figure}{0}
\renewcommand{\thefigure}{\arabic{figure}}

\begin{abstract}
Dexterous teleoperation requires precise arm–hand coordination, low-latency feedback, and robust interaction in real-world contact-rich environments. This paper presents a modular bilateral teleoperation framework that integrates \textit{operator-side} input interfaces with a \textit{robot-side} dexterous hand and compliant robotic arm in a unified control architecture. The system supports position-based hand retargeting, differential arm control, multi-scale haptic feedback, and shared control for stable manipulation. We validate the framework through a real-world dexterous manipulation task, highlighting coordinated arm–hand control and contact-aware interaction. Beyond feasibility, we identify key design insights related to cross-embodiment mismatch, haptic feedback granularity, and shared control. The proposed platform provides a practical teleoperation system and a foundation for collecting high-quality demonstrations for future learning-from-demonstration research.
\end{abstract}


\vspace{-1em}

\section{Introduction}
\label{sec:introduction}
Remote physical interaction is a critical enabler of telepresence, remote manipulation, and immersive collaboration, with applications spanning manufacturing, surgical assistance, and remote education. As teleoperation technologies mature and human-in-the-loop robotics advances, the ability to interact with physical environments from a distance is evolving from specialized research systems to general-purpose human–robot interaction frameworks, including remote gaming and dexterous manipulation~\cite{darvish2023teleoperation}.

In dexterous manipulation, high-precision tasks demand not only accurate motion retargeting but also low-latency feedback, intuitive control, and reliable coordination between multiple subsystems. As a result, demonstrations in real-world, contact-rich tasks, such as manipulating objects with irregular contact dynamics, are essential for validating overall system effectiveness and usability. Prior work has shown that while learning-based policies can achieve remarkable dexterity in simulation, they often fail to transfer to physical robots due to modeling inaccuracies, sensor noise, and unmodeled dynamics~\cite{andrychowicz2020learning}. Teleoperation addresses this gap by enabling real-world demonstrations while simultaneously supporting the collection of synchronized human motion and interaction data. Such teleoperated demonstrations provide high-quality, contact-rich trajectories that are difficult to synthesize in simulation and have been successfully leveraged for learning-from-demonstration and imitation learning in dexterous manipulation and humanoid control~\cite{he2024learning,he2024omnih2o,dass2024telemoma}, making teleoperation a critical bridge between human expertise and robot learning.



Despite the recognized importance of real-world evaluation, existing teleoperation systems remain limited in their ability to support full-hand control, rich haptic feedback, and seamless integration across heterogeneous devices~\cite{buckley2024dexterous,dass2024telemoma,he2024omnih2o,penco2024mixed}. Many approaches emphasize gross motion retargeting, often at the expense of haptic transparency and fine manipulation fidelity, while evaluation is frequently confined to simplified laboratory settings or simulation. 



These shortcomings restrict adaptability, scalability, and reduce user immersion and acceptance. Prior efforts such as~\cite{buckley2024dexterous} highlight the need for unified, low-latency architectures capable of integrating force feedback while supporting real-world deployment. Similarly, recent systems like ACE \cite{yang2025ace} and Open-Television~\cite{cheng2024open} further demonstrate the growing importance of cost-efficient, visually immersive teleoperation with bidirectional control in dexterous manipulation.


To address these gaps, we present a modular bilateral teleoperation framework for dexterous manipulation that integrates arm–hand motion retargeting, multi-scale haptic feedback, and shared control within a unified architecture. The framework is designed for extensibility, real-time interaction, and data collection in human-in-the-loop control. It serves both as a practical teleoperation system and as a research platform for studying system design choices and enabling future learning-from-demonstration methods.


\section{System Overview}
\label{sec:overview}

The proposed teleoperation framework couples a dexterous robotic arm–hand system with multiple human input interfaces. As shown in Fig.~\ref{fig:system_overview}, the system is organized into operator-side and robot-side stations, enabling interchangeable input interfaces to control coordinated arm–hand manipulation in the remote environment. The robot side comprises a 3-digit robotic hand mounted on a compliant robotic arm, while the operator interacts through complementary input interfaces, including a hand exoskeleton for hand control and a grounded haptic interface for arm control. These systems are orchestrated by a dedicated teleoperation framework that integrates motion retargeting, shared control, and haptic rendering, and are coordinated through a ROS-based software stack (Sec.~\ref{sec:overview:teleoperation_framework}). 


\subsection{Teleoperation Framework}
\label{sec:overview:remote_robotic_system}

\subsubsection*{Operator-side station}
\label{sec:overview:human_input_interfaces}

The human input interfaces provide motion commands and haptic feedback, enabling coordinated control of arm- and hand-level behaviors during teleoperation. Arm-level input and kinesthetic feedback are provided by the \textit{Lambda.7} grounded haptic interface, which supplies end-effector pose and gripper commands while rendering task-space forces via virtual coupling with the remote robotic arm. 
Hand-level input is captured using the \textit{Maestro} hand exoskeleton, which tracks 10 joint degrees of freedom across the thumb, index, and middle fingers using distributed analog sensing, enabling real-time estimation of human hand kinematics~\cite{zhang2025human}. Furthermore, the exoskeleton is equipped with series elastic actuation that supports digit-level force feedback to the user~\cite{yun2017maestro,deshpande2020robotic}.

In addition to the primary interfaces, secondary inputs such as a 3D SpaceMouse and keyboard are supported for teleoperation and mode switching, as illustrated in Fig.~\ref{fig:system_overview}. Together, these interfaces integrate into the teleoperation pipeline to provide synchronized arm and hand inputs while supporting multi-scale kinesthetic haptic feedback critical for dexterous manipulation.

\subsubsection*{Robot-side station}
The remote robotic system provides the manipulation capabilities of the teleoperation loop through a compliant arm--hand platform. It consists of the \textit{Optimo} arm (Roboligent Inc., Round Rock, TX), a commercial 7-DoF series-elastic actuator (SEA) arm, and the \textit{Aristo} dexterous robotic hand, which will be referred to as the \textit{Plato} hand, mounted at its end effector \cite{kang2026platohand,kim2026aristo}. While arm-level elasticity enables force sensing and interaction modulation for stable contact-rich manipulation, the robotic hand provides multi-finger actuation and fingertip-level force sensing for interaction with the environment. A Cartesian impedance controller computes operational-space control for the arm--hand system, impedance behavior, and joint/torque limits. Together, the arm--hand system executes retargeted and shared-control commands generated by the teleoperation framework, serving as the physical embodiment of the remote manipulation task.

\subsection{Teleoperation Framework}
\label{sec:overview:teleoperation_framework}

The teleoperation control framework connects human input to robot actions and sends contact forces back to the operator. It comprises hand tracking, motion retargeting, haptic feedback, shared control, and safety constraints. A detailed control diagram is provided in Supplementary Materials.

\subsubsection*{Hand and Wrist Tracking}
Human finger kinematics are estimated from the \textit{Maestro} exoskeleton using joint-level analog sensing and a calibrated kinematic model of the hand--exoskeleton interface~\cite{zhang2025human}. A subject-specific calibration aligns exoskeleton joints with the user’s anatomical hand configuration, enabling real-time joint-state estimation used by the retargeting and haptic modules. On the other side, wrist-level input is obtained from the \textit{Lambda.7} device as differential motion relative to an equilibrium point and drives the robotic arm task-space controller.

\subsubsection*{Motion Retargeting}
Joint angles estimated from \textit{Maestro} are mapped to \textit{Plato} through a position-based joint-to-joint mapping, enabling direct finger motion transfer between the human and robotic hands (see Fig.~\ref{fig:demonstration}(a)). To improve controllability across mismatching embodiments, the hand retargeting combines joint-to-joint mappings with spline interpolation and a unified cost function that maps human joint configurations to robot actuation, following optimization-based retargeting approaches~\cite{lakshmipathy2025kinematic,xin2026analyzing}.  

In contrast to hand retargeting, arm-level teleoperation is formulated in differential space. Wrist poses from the \textit{Lambda.7} interface are interpreted as velocity-level inputs relative to an equilibrium configuration and converted into task-space end-effector motions via inverse kinematics and Cartesian control (see Fig.~\ref{fig:demonstration}(b)). This differential formulation accommodates differences in workspace scale and kinematic structure between the human operator and the robotic arm, enabling stable and coordinated arm--hand teleoperation.

\subsubsection*{Haptic Feedback}
Haptic feedback is rendered at both hand and arm levels through the \textit{Maestro} hand exoskeleton and the \textit{Lambda.7} grounded haptic interface. Together, hand- and arm-level haptic feedback provide complementary information about local contact events and global interaction forces during teleoperation.

At the hand level, \textit{Maestro} implements high-fidelity torque control via series elastic actuators, enabling active transparency and torque-, stiffness-, and damping-based kinesthetic feedback at joint and digit levels~\cite{yun2016accurate,yun2017maestro}. Furthermore, a contact-detection module leverages sensory information from the robotic hand (fingertip force measurements), which are mapped to joint torques and rendered as discrete haptic cues indicating contact onset and/or object slip.

At the arm level, interaction forces estimated at the robotic end-effector are mapped to task-space forces and rendered through the \textit{Lambda.7} interface to guide the user’s hand during constrained maneuvers. 

\subsubsection*{Shared Control}

Shared control combines user commands with reactive assistance, including grasp stabilization, force limits, joint-level collision-aware impedance on \textit{Plato} hand, and virtual task-space constraints for \textit{Optimo} Arm, to improve robustness and safety during contact-rich manipulation while preserving operator intent.

At the hand level, shared control stabilizes grasp execution by assisting the user during contact transitions. Once an object is grasped, a fixed-trajectory impedance controller runs in parallel with manual retargeting to maintain contact and prevent slippage without overriding user motion commands.

At the arm level, shared control is implemented through virtual constraints and force limits that bound motion to avoid collisions with workspace objects while preserving compliant interaction. The \textit{Lambda.7} interface provides wrist-level pose commands by specifying end-effector poses in task space, while inverse kinematics resolves joint motion on the robotic arm. During operation, the arm yields compliantly to contact forces up to a defined safety threshold and suppresses commands that would exceed allowable interaction forces.

\subsubsection*{Implementation and Integration}
The teleoperation framework is implemented using a ROS~2-based software stack that enables real-time communication among sensing, control, and actuation components. Standardized interfaces support modular device integration and allow independent development, testing, and debugging of teleoperation modules. Additional details on the communication layers and software architecture are presented in the Supplementary Materials.

\begin{figure*}[t]
    \centering
    \includegraphics[width=\textwidth]{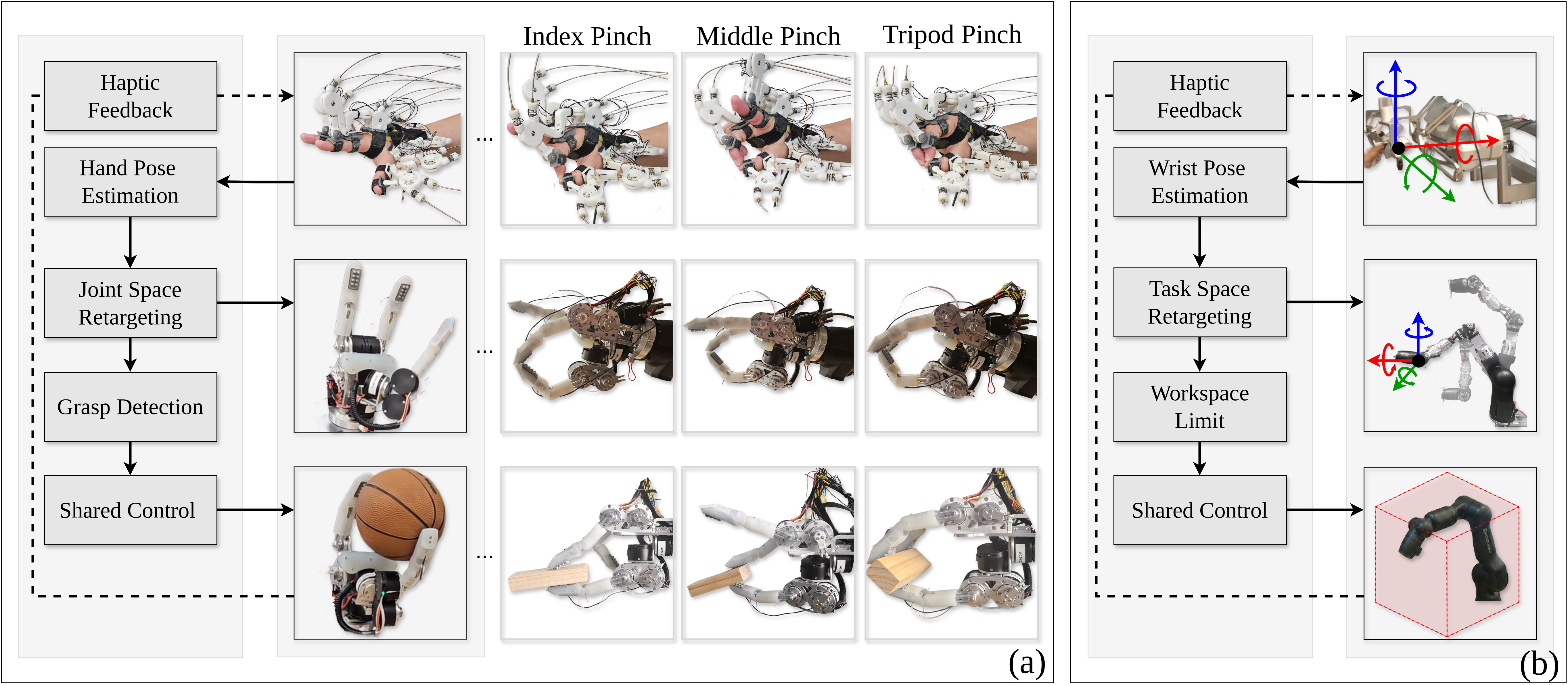}
        \setcounter{figure}{1} 
    \caption{
    Control schemes and real-world demonstrations of the proposed teleoperation strategy.
    The figure illustrates both hand-level (a) and arm-level (b) teleoperation, including the control pipeline and exemplary tasks demonstrating dexterous grasping and coordinated arm motion.
    }
    \label{fig:demonstration}
\end{figure*}

\section{Real-World Demonstration}
\label{sec:demonstration}

To evaluate the integration and feasibility of the proposed teleoperation framework, we conducted a real-world demonstration centered on the game of \textit{Jenga}. This task provides a representative benchmark for dexterous telemanipulation, requiring precise coordination of probing, grasping, contact modulation, and object placement under real-time human control. The hand-level teleoperation was performed using the \textit{Maestro} exoskeleton to control the \textit{Plato} robotic hand~\cite{kang2026platohand}. Fig.~\ref{fig:demonstration}(a) shows representative human hand poses alongside the corresponding robotic grasps. The operator successfully executed multiple grasp types, including pinch and lateral grasps. Instead, the arm was controlled using the \textit{Lambda.7} grounded haptic interface, which provided end-effector pose commands and task-space force feedback. Fig.~\ref{fig:demonstration}(b) illustrates example wrist motions and the corresponding robotic arm configurations. The operator accurately positioned the end effector around the tower and executed constrained approach and retraction motions while maintaining compliant interaction.

The complete \textit{Jenga} task consisted of three phases: probing candidate blocks, extracting a selected block, and placing it on top of the tower. Throughout the task, the system maintained stable coordination between hand and arm control, enabling successful execution without toppling the tower. Across repeated trials, the framework consistently supported successful block extraction and placement while maintaining stable bilateral operation. Representative demonstrations are provided in the Supplementary Video.

\section{Discussion}

We presented a modular bilateral teleoperation framework for real-world, contact-rich dexterous manipulation with coordinated arm–hand control. Hardware demonstrations confirm the feasibility of bidirectional teleoperation with integrated motion retargeting, haptic feedback, and shared control.

Beyond feasibility, the system provides a research platform for studying hand tracking, retargeting, haptic feedback, and shared control in human-in-the-loop manipulation, and for collecting demonstrations to support future learning-from-demonstration pipelines~\cite{qin2022one,si2024tilde}.

Overall, our experiments highlight that effective dexterous teleoperation depends less on maximizing hardware fidelity and more on user-centered system integration. In particular, robustness to human variability, appropriate abstraction of haptic feedback, morphology-aware retargeting, and non-autonomous shared control emerged as critical design dimensions. 



\subsection{Hand Morphology and Donning Effects}

Wearable hand interfaces are highly sensitive to user anatomy and donning conditions, which introduce systematic errors in joint estimation and degrade retargeting fidelity~\cite{xia2022hand,yousaf2023experimental}. In our experiments, even minor variations in hand size or device alignment produced noticeable errors in estimated hand kinematics, directly affecting teleoperation precision.

When hand-size invariance cannot be achieved through mechanical or sensing design alone, explicit calibration or adaptive tracking becomes necessary. Because the kinematic mapping of the \textit{Maestro} exoskeleton is sensitive to donning conditions, we therefore employ a custom calibration procedure to align the sensed exoskeleton joint space with the user’s anatomical hand configuration~\cite{zhang2025human}.

Overall, our observations motivate morphology-aware and adaptive hand tracking pipelines that preserve fidelity across users and donning conditions while minimizing setup burden.

\subsection{Impact of Haptic Feedback}

Haptic feedback in dexterous teleoperation can be rendered at different levels of granularity, ranging from joint-by-joint force rendering to sparse, contact-based cues~\cite{pacchierotti2017wearable}. The effectiveness of haptic feedback depends strongly on this granularity and the task context rather than force fidelity alone~\cite{rodriguez2024measuring}. Highly detailed feedback may overwhelm operators and increase cognitive load, whereas overly sparse feedback may fail to convey task-relevant information, particularly in contact-rich manipulation.

Our system supports joint-, digit-, and hand-level haptic feedback evaluated across multiple teleoperation modes. Digit-level contact cues derived from robotic hand sensing were reported by operators to improve perception of contact onset and slippage and were preferred over purely visual feedback in precision tasks, consistent with prior findings on reduced workload and improved operability~\cite{tanioka2025effects}. In contrast, the highest-granularity feedback occasionally overwhelmed operators, depending on the implementation and task demands.

Overall, these observations reinforce that greater hardware or rendering capability does not necessarily yield better user experience. Instead, effective haptic feedback should be designed and evaluated from a user-centered perspective, prioritizing perceived control, cognitive load, and task performance over hardware-centric metrics~\cite{rodriguez2024measuring,hidalgo2025evaluating}.

\subsection{Motion Retargeting for Cross-Embodiment Mismatch}

Teleoperation across heterogeneous embodiments inherently introduces kinematic and morphological mismatches that limit the effectiveness of direct joint-level mappings. In our system, differences between the human hand, the \textit{Maestro}-derived hand model, and the \textit{Plato} robotic hand manifested as reduced controllability when strict joint correspondence was enforced. Our observations support recent findings that effective retargeting should be framed around task-level objectives, such as fingertip placement, contact stability, and grasp feasibility, rather than exact joint matching, particularly under morphological mismatch~\cite{meattini2022human,xin2026analyzing}.

More broadly, our results suggest that emphasizing task-relevant contact behavior while enforcing feasible joint configurations is critical for achieving intuitive and stable teleoperation across embodiment differences. This motivates morphology-aware retargeting frameworks that adapt objective weighting to both embodiment and task context, rather than relying on fixed kinematic mappings, in line with emerging trends in dexterous teleoperation research~\cite{wen2025dexterous}.

\subsection{Shared Control for Human Error Mitigation}

Teleoperation inevitably involves residual errors arising from imperfect hand tracking, retargeting inaccuracies, latency, and human variability. Our experiments confirmed that even with careful system design, these errors cannot be fully eliminated and place a continuous cognitive burden on the operator. A key lesson learned is that shared control can meaningfully mitigate residual human error without requiring full autonomy. 

By enhancing teleoperation commands with task-relevant constraints, the system can improve robustness during contact-rich manipulation while preserving operator intent.
Prior work similarly reports that hybrid shared-control strategies improve stability and task performance by balancing human input with safety- and task-oriented constraints~\cite{nicolis2020general,selvaggio2019passive,xi2019robotic}. Our observations highlight that even simple, non-autonomous shared-control mechanisms can substantially enhance dexterous teleoperation by reducing error sensitivity while maintaining direct human control.

\section{Conclusion}
This paper presented a modular bilateral teleoperation framework for contact-rich dexterous manipulation and demonstrated its feasibility on real hardware. The main lesson is that  usable dexterous teleoperation depends less on maximizing hardware fidelity and more on morphology-aware retargeting, task-relevant haptic abstraction, and lightweight shared-control mechanisms that stabilize interaction while preserving operator intent.

The proposed system provides a practical platform for studying human-in-the-loop dexterous manipulation and for collecting high-quality demonstrations to support future learning-from-demonstration and adaptive shared-control methods.

\bibliographystyle{ieeetr}
\bibliography{references}

\end{document}


\begin{table*}[t]
\caption{Descriptions of supplementary materials}
\label{tab:supplementary}
\centering
\begin{tabularx}{\textwidth}{lX}
\toprule
\textbf{Supplementary material} & \textbf{Description} \\
\midrule
Project website &
Project websites hosting videos, documentation and other supplementary material for the devices used in the bilateral teleoperation framework. \\[1em]
& \textbf{Plato}:~\url{https://platohand.github.io/}\\
& \textbf{Aristo}:~\url{https://aristohand.github.io/}\\
& \textbf{Maestro}:~\url{https://reneu.robotics.utexas.edu/research/maestro-teleoperation}\\[1em]

Videos &
Real-world demonstrations of the teleoperation framework performing the Jenga task and additional dexterous manipulation trials, highlighting coordinated arm–hand control and shared control behaviors. \\[2em]

Appendix &
Additional technical details on the teleoperation framework, including control logic diagram, communication architecture and timing, and hardware implementation aspects that complement the main text. \\[2em]

\bottomrule
\end{tabularx}
\end{table*}

\clearpage

\title{Appendix}

\maketitle

\begin{figure*}[t]
    \centering

    \begin{subfigure}[t]{0.24\textwidth}
        \centering
        \includegraphics[width=\linewidth]{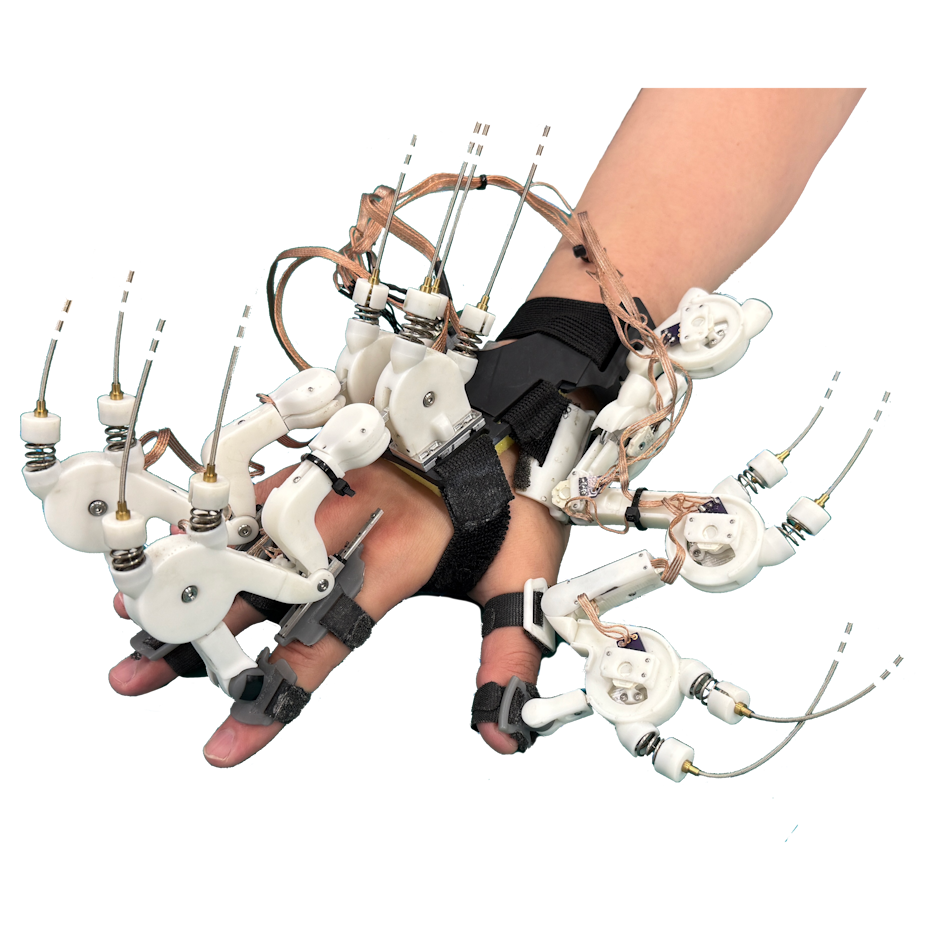}
        \caption{Maestro Hand Exoskeleton}
        \label{fig:appendix_maestro}
    \end{subfigure}
    \hfill
    \begin{subfigure}[t]{0.24\textwidth}
        \centering
        \includegraphics[width=\linewidth]{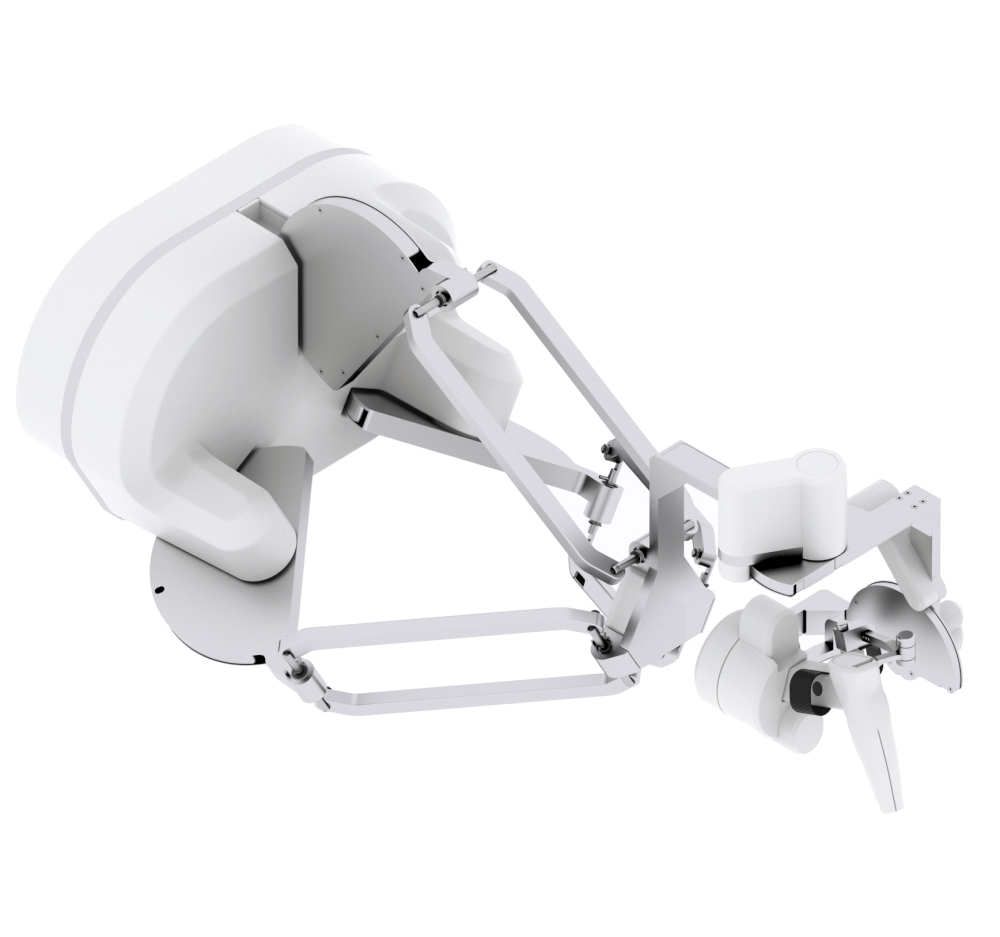}
        \caption{Lambda.7 Haptic Interface}
        \label{fig:appendix_lambda}
    \end{subfigure}
    \hfill
    \begin{subfigure}[t]{0.24\textwidth}
        \centering
        \includegraphics[width=\linewidth]{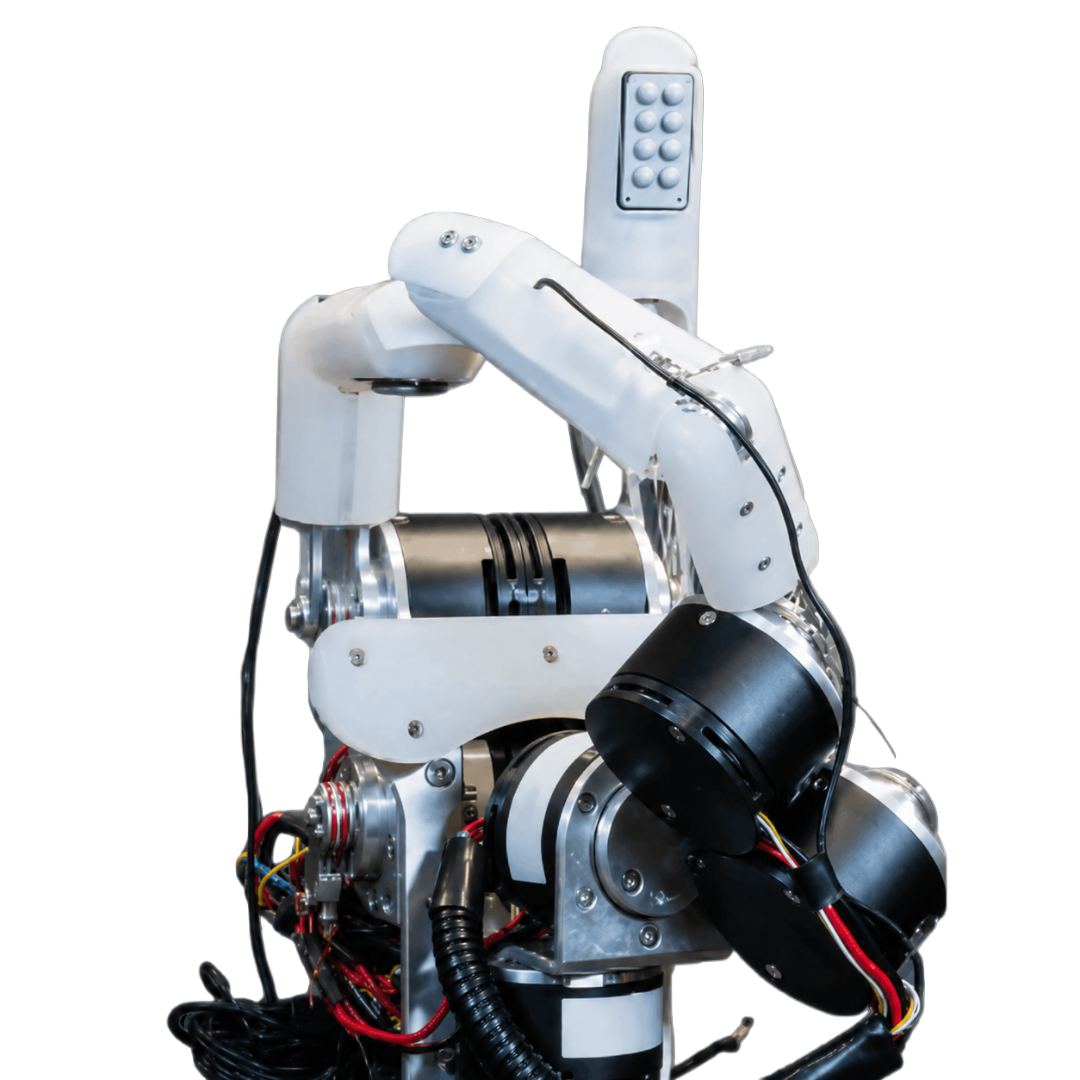}
        \caption{Plato Robotic Hand}
        \label{fig:appendix_plato}
    \end{subfigure}
    \hfill
    \begin{subfigure}[t]{0.24\textwidth}
        \centering
        \includegraphics[width=\linewidth]{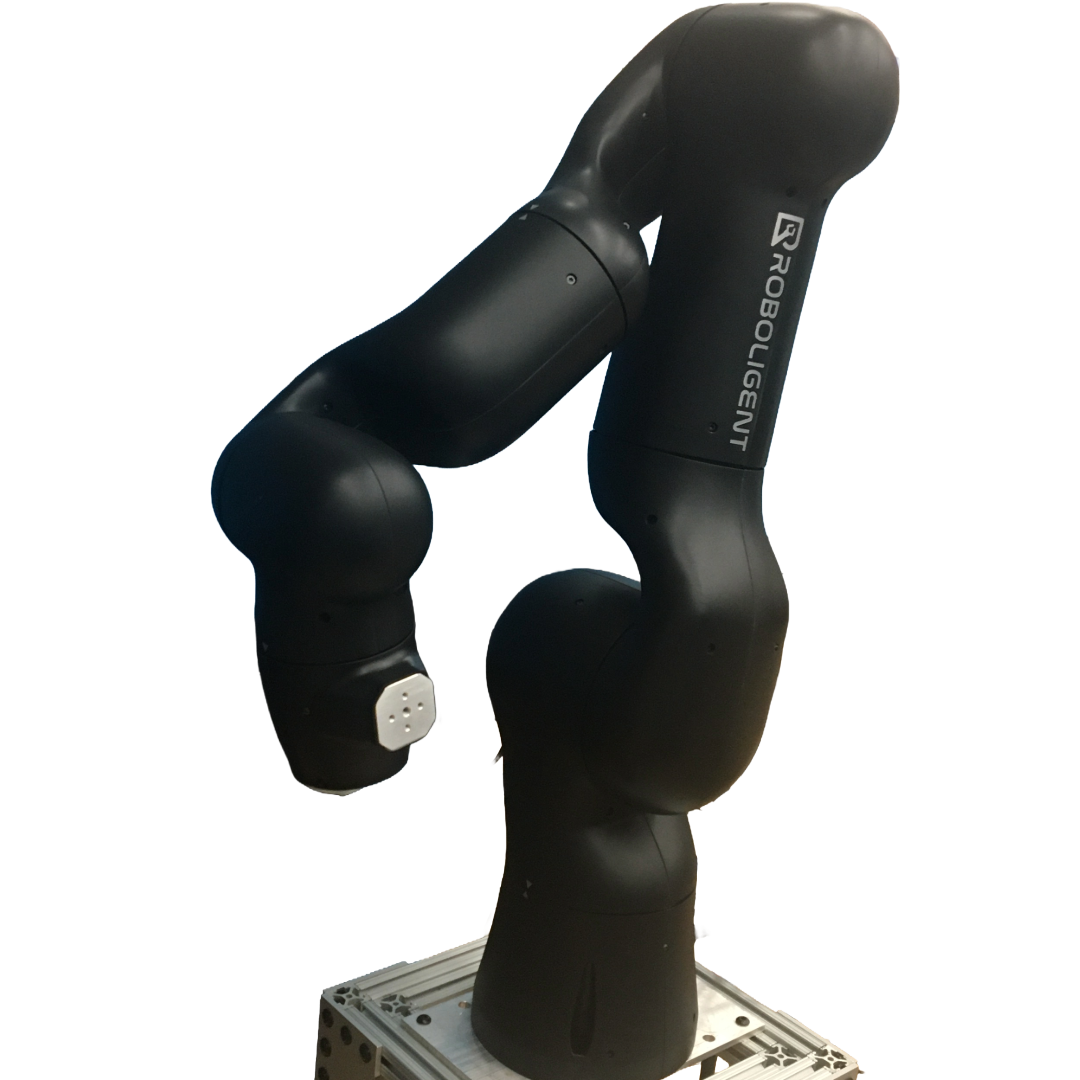}
        \caption{Optimo Robotic Arm}
        \label{fig:appendix_optimo}
    \end{subfigure}

        \caption{Hardware components of the presented bilateral teleoperation framework. The operator-side station consists of the  \textit{Maestro} hand exoskeleton and the \textit{Lambda.7} haptic interface, while the robot-side station comprises the \textit{Plato} robotic hand and the \textit{Optimo} robotic arm.}
    \label{fig:appendix_hardware}
\end{figure*}

    \begin{figure*}[t]
        \centering
        \includegraphics[width=\linewidth]{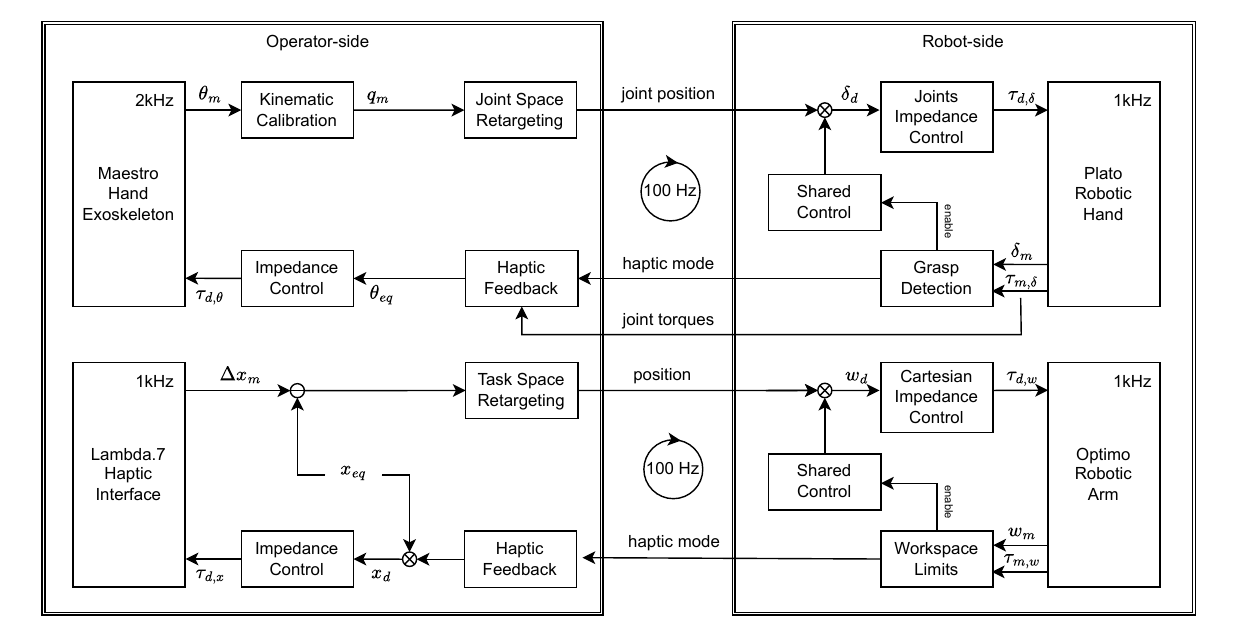}
        \caption{Complete control diagram of the teleoperation architecture, detailing signal flow between the operator-side interfaces and the robot-side arm–hand system, including the associated low-level control schemes.}
        \label{fig:appendix_control}
    \end{figure*}

\subsection{Maestro: Hand Exoskeleton}

The \textit{Maestro} hand exoskeleton (Fig.~\ref{fig:appendix_maestro}) serves as the primary hand-level interface for capturing operator finger motion and rendering digit-level haptic feedback derived from robotic hand interactions. Additional details on its mechanical design and characterization are provided in~\cite{yun2017maestro}.

It tracks the thumb, index, and middle fingers using distributed analog sensing: 16 rotary potentiometers measure joint motion, with redundant sensing on each digit to support joint and fingertip estimation~\cite{zhang2025human}, and provides joint-level haptic feedback through a SEA-based actuation mechanism~\cite{yun2016accurate}. Low-level joint torque control is implemented via a velocity-sourced torque loop that uses estimated joint torques from series elastic actuator displacements, while mid-level joint control is executed through explicit impedance control.
Haptic feedback is rendered through an impedance-control scheme in which the equilibrium configuration is set to the hand pose at contact onset. The controller stiffness is manually tunable to accommodate user preferences and to adjust the responsiveness of the system. 

The mechatronic system uses custom EtherCAT-based motor drivers (EsmaCAT, Austin, TX), with the EtherCAT master running on a PREEMPT-RT Linux real-time machine. Low-level control loops execute at 2~kHz on a dedicated real-time machine, while high-level communication with the teleoperation framework is handled through a custom driver and ROS~2 bridge. Desired joint commands and haptic cues are exchanged at 100~Hz, together with joint states, estimated torques, and device status information used for retargeting, shared control, and safety monitoring.

\subsection{Plato: Robotic Hand}

The \textit{Plato} robotic hand (Fig.~\ref{fig:appendix_plato}), which is also known as Aristo, comprises a three-fingered, 8-DoF kinematic structure \cite{kang2026platohand, kim2026aristo}. During teleoperation, the \textit{Plato} hand executes retargeted finger motions generated from the operator's hand movements and provides fingertip contact information used for haptic feedback and shared-control behaviors. As such, it serves as the primary manipulation and contact-sensing component of the robot-side station.

The system articulates the index and middle fingers via 2-DoF flexion-extension joints and employs a 4-DoF joint for the thumb. 1:8 planetary geared Quasi-Direct Drive (QDD) actuators drive all joints to enable proprioceptive force estimation, while pre-tensioned cables transmit a wide range of motion to the distal joints. An embedded driver executes a 1~kHz impedance control loop that regulates torque, position, velocity, stiffness, and damping parameters.

The high-level software operates within the ROS~2 Control framework and manages the hardware through a custom impedance controller. The host PC updates command setpoints to the embedded drivers via CAN bus at 100~Hz. Separately, NARI-Touch tactile sensors at each fingertip transmit contact data directly to the PC via USB at 100~Hz.

\subsection{Optimo: Robotic Arm}

The \textit{Optimo} robotic arm (Fig.~\ref{fig:appendix_optimo}) is a commercial 7-DoF force-controlled robotic manipulator manufactured by Roboligent (Round Rock, TX, USA), with a 6 kg payload capacity and a 1100 mm reach. Each joint is actuated through a series elastic actuator, enabling compliant interaction and joint-level torque sensing. The arm communicates through EtherCAT for real-time multi-axis control and provides a joint torque control resolution of 1.5 mNm at the actuator output and an end-effector force control resolution of 0.025 N. 

During teleoperation, the \textit{Optimo} arm serves as the robotic embodiment of the operator's arm motions and provides the mounting platform for the \textit{Plato} hand. Its compliant actuation enables stable interaction with the environment while executing teleoperation commands. Combined with its redundant kinematic structure and high-resolution force control, this makes the arm well suited for force-sensitive teleoperation and contact-rich manipulation.

\subsection{Lambda.7: Haptic Interface}
The \textit{Lambda.7} (Fig.~\ref{fig:appendix_lambda}) is a 7-DoF haptic feedback end-effector device (Force Dimension, Nyon, Switzerland), which provides high-fidelity translation, rotation, and gripper feedback. The haptic controller runs at a 1-kHz feedback rate via ROS~2 enabling rendering of stiff and compliant interactions. The gripper optionally serves as a universal action button enabling discrete predefined grasps to be commanded to the \textit{Plato} hand.

During teleoperation, the device provides differential end-effector pose commands relative to an equilibrium configuration. Interaction forces estimated at the robot end-effector are mapped to task-space forces and rendered through the device using a virtual coupling scheme. This enables the operator to perceive large-scale interaction forces while maintaining stable bilateral control.

\subsection{Software Architecture and Timing}

The teleoperation framework is implemented using ROS 2 and
coordinates communication between operator-side and robot-side
devices. High-level teleoperation commands are exchanged at
100 Hz, while device-specific control loops run at their native
rates. \textit{Maestro} executes low-level control at 2 kHz, \textit{Plato} and \textit{Lambda.7} operate at 1 kHz, and tactile sensing is streamed at
100 Hz. Fig. \ref{fig:appendix_control} summarizes the complete signal flow and timing relationships among subsystems.

\bibliographystyle{ieeetr}
\bibliography{references}